\documentclass[11pt]{article}

\usepackage[a4paper,left=1.1in,right=1.1in,top=1.1in,bottom=1.1in]{geometry}
\usepackage{graphicx} 

\usepackage{hyperref}       
\PassOptionsToPackage{hyphens}{url}
\usepackage{url}            
\usepackage{booktabs}       
\usepackage{amsfonts}       
\usepackage{amsmath}
\usepackage{cleveref}
\usepackage{nicefrac}       
\usepackage{microtype}      
\usepackage{graphicx} 
\usepackage{wrapfig}
\usepackage{enumerate}
\usepackage{tikz,tikz-qtree}
\usepackage{import}
\usepackage{enumitem}
\usepackage{subcaption}
\usepackage[numbers]{natbib}
\usepackage{comment}
\usepackage{wrapfig}
\usepackage[T1]{fontenc}

\usepackage[textsize=scriptsize, textwidth=2.5cm]{todonotes}

\usepackage{xcolor}         
\hypersetup{
    colorlinks=true,    
    urlcolor=blue,
    }

\usepackage{twemojis}

\NewDocumentCommand\emojigpt{}{
    \includegraphics[scale=0.01]{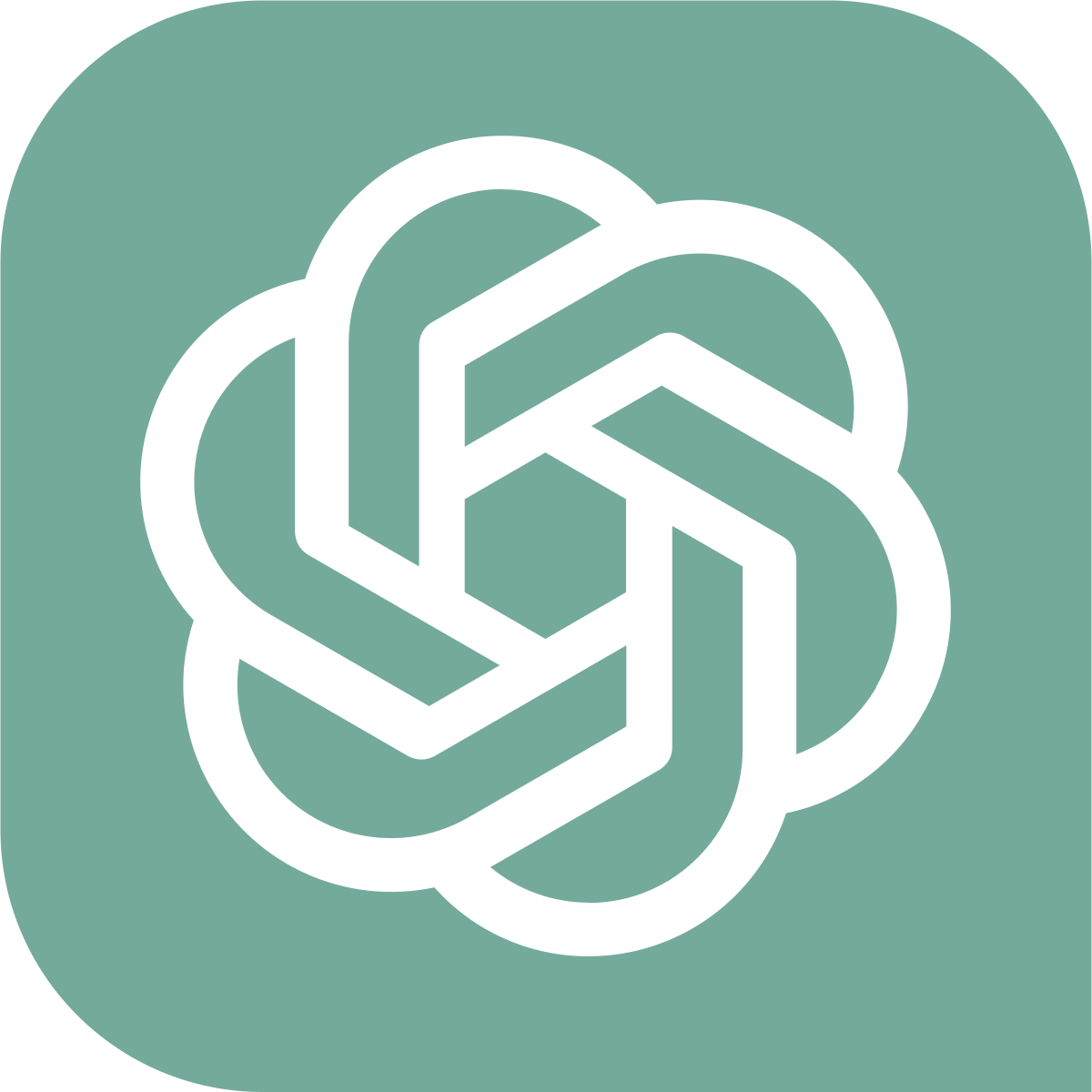}
}

\usepackage{tcolorbox}
\definecolor{lightblue}{HTML}{18282e}
\definecolor{lighterblue}{HTML}{f2fafd}  
\newtcolorbox{abox}{colback=lighterblue,colframe=lightblue}

\usepackage{parskip}

\usepackage{xpatch}
\makeatletter
\xpatchcmd{\@maketitle}
  {center}
  {flushleft}
  {}{}
\xpatchcmd{\@maketitle}
  {center}
  {flushleft}
  {}{}
\xpatchcmd{\@maketitle}
  {{c}}
  {{@{}l@{}}}
  {}{}
\xpatchcmd{\and}
  {{c}}
  {{@{}l@{}}}
  {}{}
\makeatother

\title{Perspectives on the State and Future of Deep Learning - 2023}
\date{ }
\author{\textbf{Editors:} Micah Goldblum\footnote{\href{goldblum@nyu.edu}{goldblum@nyu.edu}}, Anima Anandkumar, Richard Baraniuk, Tom Goldstein\footnote{\href{tomg@umd.edu}{tomg@umd.edu}}
\\ \textbf{Contributors:} Kyunghyun Cho, Zachary C Lipton, Melanie Mitchell,\\ Preetum Nakkiran, Max Welling, Andrew Gordon Wilson}

\begin{document}

\setlength{\twemojiDefaultHeight}{11pt}

\maketitle

\begin{abstract}
\normalsize 

The goal of this series is to chronicle opinions and issues in the field of machine learning as they stand today and as they change over time.  The plan is to host this survey periodically until the AI singularity paperclip-frenzy-driven doomsday, keeping an updated list of topical questions and interviewing new community members for each edition. In this issue, we probed people's opinions on interpretable AI, the value of benchmarking in modern NLP, the state of progress towards understanding deep learning, and the future of academia.
\end{abstract}

{  \hypersetup{linkcolor=blue}
\tableofcontents
}

\vspace{5mm}
\phantomsection
\addcontentsline{toc}{section}{What is the biggest problem that we should be working on but aren’t?}
\begin{abox}
\textbf{What is the biggest problem that we should be working on but aren’t, or vice versa?} What other areas will see progress other than text and images, and how might this impact fields and industries outside computer science?
\end{abox}

\vspace{3mm}

\textbf{Max Welling:} I am a huge fan of applying AI to the natural sciences. There are an enormous number of very exciting and impactful applications in this area. For example, we can predict global warming using fast PDE surrogates, we can design new battery and carbon capture materials through faster density functional theory and molecular dynamics simulators and generative AI, and we can predict the effectiveness of new drugs using these methods. The natural sciences provide a treasure trove of data and exciting problems to work on!
\vspace{3mm}

\textbf{Zachary Lipton:} It’s hard to guess what problems are truly being neglected at this strange moment in history when seemingly everyone in every walk of life, rightly or wrongly, is looking to apply AI to their problems. In short, I think this ship has sailed. AI methods are already applied far outside of computer science, and even in industry, it’s no longer strictly scientists at the helm the way it was, say, 5 years ago. I’m surprised constantly by how many business people are following AI so closely, by how often they are abreast of new developments before academic researchers and by how academia’s influence has waned in many corners of industry. Of course there are always profound problems to address, and the hard problems are competing for the attention of a relatively small set of extreme athletes. I have my own passion problems—tackling the documentation at the root of physician burnout, leveraging data to realize the dream of personalized medicine—but while I’d welcome more hands on deck, you can hardly claim that nobody’s working on them.
\vspace{3mm}

\textbf{Kyunghyun Cho:} I have two things in my mind. First, we seem to focus way too much on simple problems where answers/predictions are readily verifiable by e.g. simple surface-level comparison (0-1 loss against the gold standard answer) or by asking a crowd worker. These problems are also the ones that have the most amount of data available. In other words, on these simple problems, we can iterate rapidly due to relatively cheap evaluation, and our models work quite well from the beginning due to the abundance of data. I believe we should consider more challenging problems where each iteration of development is costly, which will force us to look into more interesting and challenging topics within machine learning and more broadly artificial intelligence, such as active learning and learned evaluation.

Second, I believe we should work on so-called distribution shifts much more so and much more carefully. A lot of people work on distribution shifts, but are often working in a more traditional mode of empirical research in machine learning; that is, they build a benchmark and iterate on developing algorithms based on the accuracy on the benchmark. Unfortunately this paradigm is completely broken for studying distribution shifts, as the designer can use their knowledge of the test distribution when choosing the algorithm. By computing accuracies or even inspecting data a bit, distributional information of the test set leaks, and this largely nullifies various findings. Rather, we have to always start from specifying what kind of distribution shifts we are considering, and come up with an algorithm prior to testing it on any benchmark.

Both of these directions will greatly improve the applicability and impact of machine learning. As for me personally, active learning and robust classification are two key problems I am seeing every day at Prescient Design, Genentech for antibody design.
\vspace{3mm}

\textbf{Melanie Mitchell:} I will list a few “biggest problems” that we should be working on (and indeed some people are): 

\begin{enumerate}
\item Reasoning about abstract concepts (e.g., Chollet’s ARC challenge \cite{chollet2019ARC}):  This is a fundamental human ability that AI systems have not yet mastered in any general way.  For example, humans are able to solve (at least a large percentage of) problems in Chollet’s “Abstraction and Reasoning Corpus,” \cite{ARC-Github} which tests for few-shot abstraction abilities and general understanding of “core concepts”.  No AI system today comes close. 

\item Multimodal common sense (e.g., compositionality, visual reasoning):  Current multimodal vision-language models (e.g., CLIP, DALL-E, etc.) are known to be poor at compositional understanding, reasoning, and general visual common sense (e.g., they are susceptible to non-humanlike errors and adversarial attacks).  Dealing with the integration of vision and language, and the physical and social understanding these require, are of course essential capabilities for any AI system interacting with the human visual world.  

\item Better ways of evaluating AI systems’ capabilities:  The most common methods of evaluation—static benchmarks, reporting only aggregate statistics such as test-set-accuracy—do not give enough insight into the actual capabilities and robustness of AI systems.  Moreover, giving AI systems tests designed for humans (e.g., the Bar exam, medical licensing tests, IQ tests, etc.) is not always a good way of evaluating their capabilities, since those tests make many assumptions that are true for humans but not for machines.  New ways of evaluating the abilities and generality of AI systems are needed.  One promising approach is for AI researchers to collaborate with cognitive scientists—such as developmental psychologists and animal cognition researchers—who have experience in designing controlled experiments to gain insight into “alien intelligences” (cf. \cite{frank2023baby})

\item Mechanistic interpretability:  This refers to understanding how AI systems work “under the hood”—that is, the actual mechanisms underlying their information processing.  Just as mechanistic interpretability is an unsolved problem in brains, it’s a very difficult and unsolved problem for today’s large-scale pretrained transformer-based models.  However, understanding underlying mechanisms of behavior is a central goal of any science, and it would be extremely useful to have a well-developed science of AI that attempts to find out what these mechanisms are. I believe this would give us insight into the capabilities and limitations of AI systems that we can’t discover from behavioral observations alone.
\end{enumerate}

\vspace{3mm}

\textbf{Andrew Gordon Wilson:} Machine learning for scientific discovery. There are many efforts applying machine learning to scientific tasks — even under the headline of ``AI for Science'', a phrase often used in recruiting. Yet there are relatively few efforts developing machine learning to discover fundamental scientific insights into data that we didn't have before, towards new scientific theories. Moreover, the unifying theory — the general understanding — rather than any particular application, has historically been the primary quantity of scientific interest! 

Consider general relativity (GR) as a running example. GR is primarily exciting not because it enables GPS, which is an impactful application, but because it tells us how gravity, time, and space interact, towards a multitude of different applications. GPS wasn't even envisaged when GR was proposed. While we probably could hack a neural network together to correct for gravitational time dilation to make GPS possible, we would lack an underlying understanding that enables many other applications, including applications we don't currently foresee. Similarly, special relativity is in some sense “merely” an interpretation of the Lorentz transformations (which were experimental correction factors), but is viewed as a much more significant contribution than the equations, due to the fundamental understanding it provides.

We worry about “AGI”, but we haven't even made the most basic steps towards a system that could propose a theory anything like general relativity or quantum mechanics. \emph{We haven't scratched the surface of scientific discovery}. What would an algorithm that could propose such theories even look like? GR was not “data driven”, though it explains some mysterious empirical observations of the time, such as Mercury's perihelion. Such an algorithm wouldn't be at all like the deep learning of today, or classical symbolic systems. Einstein wasn't enumerating symbolic expressions and selecting for ones with “nice” properties such as symmetries. 

I want to live in a future where AI profoundly advances scientific understanding --- science that we ourselves may not be capable of discovering, such as an understanding of how the brain works.

An “up and coming” direction I am personally very excited about is in exploiting algebraic structure that broadly arises as a consequence of common modelling assumptions. I don't think there is enough general awareness of how important numerical linear algebra will be as a foundation for machine learning. We have a project, and library, CoLA (Compositional Linear Algebra) that is all about this \citep{potapczynski2023cola}. In some sense, there is an analogy with autograd — automatic differentiation in libraries like PyTorch, JAX, and others before it, have revolutionized deep learning by freeing users to rapidly prototype ideas without manually computing derivatives. A similar result will be achieved with numerical linear algebra. Arguably, machine learning is essentially an application of linear algebra.

In terms of what we are working on but shouldn't, I would only say there is too much of a herd mentality in research — too many people working on essentially the same thing! For a time, this was GANs. Chasing topical research provides short term validation and attention, but it's not typically a good way to make meaningful long-term contributions, or to find work that resonates with you personally. If you're too worried about being scooped, then it's a good indication to work on something else. We tend to erroneously assume that the foundations are “set in stone”; but there are many fundamental considerations that deep learning researchers aren't even thinking of questioning.
\vspace{3mm}

\phantomsection
\addcontentsline{toc}{section}{Why haven’t we made progress towards understanding deep learning?}
\begin{abox}
\textbf{Why haven’t we made progress towards understanding deep learning and will we ever?}  If we do ever develop a theory of deep learning, would it be useful to practitioners and engineers?
\end{abox}

\vspace{3mm}

\textbf{Alex Smola:} Understanding nonparametric nonlinear functions in general is mission impossible, but so is ``understanding'' the Fourier coefficients or wavelet coefficients of a function. It's as easy or hard as properly understanding potential functions in solid state physics. 
\vspace{3mm}

\textbf{Max Welling:} I am sure we will make progress in visualizing the inner workings of DL, and to some level be able to interrogate the way it reasons. However, we may have to live with the fact that these systems reason and think in fundamental new ways that are very hard for us mere mortals to understand. For certain applications this will not be acceptable (such as decisions that impact people’s lives) and for others it may be fine (such as predicting the stock market). I am personally not too worried about the need to understand every last bit of these models as long as we can find ways to certify them for the domains that we want to use them for. After all, I do step into an airplane without a clue of how its software works.

\vspace{3mm}

\textbf{Kyunghyun Cho:}  I’m not entirely sure if we really haven’t made progress toward understanding deep learning. What do we not understand about deep learning?

A lot of techniques we use in order to build, train and deploy deep learning models and systems were invented and developed as part of understanding challenges in training deep learning models. For instance, the issue of vanishing gradients was identified in early 90’s, studied quite extensively by Yoshua Bengio and Sepp Hochreiter over the entire decade in 90’s, and we know how to address this issue (to a certain degree) by introducing linear shortcut connections, such as residual connections. 

Our understanding of loss functions typically used for training a deep neural net has only improved over the past several decades. With a particular type of deep neural nets, we know that there is no point minimum but a long and perhaps highly curved low-loss valley, and in some cases we can also show this quite rigorously, although intuition behind this observation was already known in 80’s (see Rumelhart et al.’s backpropagation paper.) This observation has led to some interesting practical algorithms such as the one from Andrew Gordon Wilson’s group.

So, my answer would be that we have made progress toward understanding deep learning, although there is much more progress in multiple directions to be made in the future, such as characterizing and tying up the patterns of generalization with neural net learning (architecture, optimization, data, etc.) Furthermore, I would say a lot of progress in theoretical understanding of various aspects of deep learning have already resulted in and will continue to result in algorithmic practices (including algorithms, recipes for training neural nets, etc.) that are useful for practitioners. 
\vspace{3mm}

\textbf{Andrew Gordon Wilson:} I disagree with the premise that we haven't made progress. Part of what has slowed progress is that we assume lack of progress as a premise, and then attempt to rationalize that premise. If we can acknowledge the progress that has been made, then we can make steps towards further progress. \emph{The real crisis is a failure to acknowledge progress}. 

And there has been substantial progress in understanding deep learning. Virtually everything about deep learning that has been pointed to as “mysterious” or “different” is neither particularly mysterious nor different from classical machine learning. Overparametrization, having more parameters than data points, was commonplace in nonparametric statistics (very popular just before the re-emergence of deep learning as a dominant paradigm in 2012), and it is understood that it is an elementary mistake to treat parameter counting as a proxy for generalization \citep[e.g.,][]{wilson2020bayesian}. Models with more parameters often provide better compressions of the data, leading to lower effective dimensionality, and simpler solutions that generalize better \citep{maddox2020rethinking}. We also understand intuitively why this is the case by examining loss landscapes: flat solutions occupy a much greater volume in high dimensions, and are linked to model compressibility \citep{ronny2019understanding, hinton1993keeping, hochreiter1997flat, mackay2003information}. These findings also explain double descent, which can exactly be reproduced with linear models \citep{krogh1992generalization, belkin2019reconciling, wilson2020bayesian}. We now know that stochastic optimization plays a limited role in generalization, and it is more the properties of the loss landscape that influence generalization \citep{geiping2021stochastic, chiang2022loss, izmailov2021bayesian}. We understand benign overfitting, again through analogy with nonparametric statistics --- we wish to embrace models that are highly flexible, but have inductive biases towards simple solutions. Such models can fit noise if there is a strong likelihood signal, but they will avoid high complexity solutions if they can fit the data without them \citep[e.g.,][]{wilson2020bayesian}. There has been enormous practical progress in developing Bayesian neural networks for uncertainty representation \citep[e.g.,][]{wilson2020bayesian, osawa2019practical, daxberger2021laplace}. We have non-vacuous generalization bounds for modern neural networks \citep[e.g.,][]{mcallester1998some, dziugaite2017computing, alquier2021user, zhou2018non, lotfi2022pac}, which in some cases can  even be prescriptive \citep{lotfi2022pac}, and do not necessarily require stochastic networks (contrary to common belief) \citep{goldblum2023}. We have mathematical mechanisms to encode equivariances \citep[e.g.,][]{cohen2016group, finzi2021practical}. We have a nuanced understandings of data augmentation \citep{geiping2022much}. And these are just a handful of references that I am most familiar with. There are many more. The list goes on.

This is not to say that everything in deep learning is understood. But there has been progress and we must properly acknowledge this progress in order for the field to keep moving forward. There are also features of neural networks that are relatively distinct, such as the mode connectivity in the loss landscapes \citep{garipov2018loss}.

I've heard it said ``we need a theory of deep learning''. What does that mean? Will we ever be satisfied that we have such a theory? I'm not sure many researchers would actually agree on what it means. We certainly have a relatively good understanding of many properties of neural networks that have been presented as mysterious, as outlined above.

Designing controlled experiments to provide a more fundamental understanding of deep neural networks is a particularly promising direction, since they are often hard to analyze theoretically without making unrealistic assumptions. Such efforts have been gaining momentum lately and need to be supported.
\vspace{3mm}

\textbf{Zachary Lipton:} To begin tackling the difficulty of understanding deep learning, I think it’s important that we understand what sort of animal deep learning is and what might be appropriate comparisons. We might view deep learning as offering a collection of model classes, and view ``understanding deep learning'' as analogous to understanding other model classes that were previously popular among academics and practitioners. We might make comparisons to linear models, kernel methods, etc., and bemoan the comparative lack of satisfying theory to account for deep learning’s capacity to generalize. 

But these direct comparisons between deep learning and previous approaches strike me as deeply misleading. With deep learning, we typically apply models directly on raw or lightly processed data. By comparison, the success of linear models (and even kernel methods) in complex domains have always depended on an extraordinary amount of feature engineering, for which we never possessed any satisfying theoretical account. While we might have felt satisfied as scholars that we “understood” something profound about generalization in these previous models, the skeptical view is that this was never where the action was. When I got to grad school (on the eve of the explosion of deep learning into mainstream machine learning), I interacted quite a bit with both theorists and practitioners. To hear the theorists speak, you’d think that the entire practice of machine learning lived in the space of the learning algorithm and that we were making tremendous progress. But among practitioners, in data mining and computer vision, the adage was that one great (hand-engineered) feature was worth far more than any difference in performance among algorithms. 

In short, I think deep learning couples together the learning of a classifier with the learning of a suitable representation, and this question—of what constitutes a suitable representation—has never been met with much understanding, neither in the deep learning era nor in previous eras. The difference now is that we now consider this to be the responsibility of “the algorithm” rather than externalizing it. 

As to the usefulness of a theory of deep learning, I expect that to depend heavily on what form that theory takes. One cliché form for titling theory papers goes as follows: ``Some provable bounds for \rule{1cm}{0.15mm}''. Should the reader care? That depends on precisely what quantities have been analyzed and how this analysis relates (or not) to the needs of the practitioner. 

At risk of sounding pessimistic, I would note that the theory of deep learning hasn’t caught up to the state of deep learning in 2014, certainly not enough to provide a reliable source of guidance for practitioners, or even for researchers interested in developing practical methodology. At the same time, the set of relevant practices has shifted dramatically. Now, we develop models for new tasks, leveraging a motley combination of specialized labeled data, human feedback, and pre-trained components drawing on web scale data resources. These kinds of settings where data is coming from multiple populations, used to fit different model components with many different objectives and combined in unpredictable ways seems especially frustrating for the pursuit of theoretical insights. 

On a positive note, I don’t think that theory must account for everything that deep learning does to be useful in deep learning settings. For example, suppose that I want to study a distribution shift problem where the underlying predictor is produced via deep learning. I may not need to explain how deep learning works to develop a coherent theory to tell me what corrective action ought to be taken to adjust in light of shifts under some set of structural assumptions. It’s often sufficient to stick the deep learning in a black box and say suppose I have a predictor with certain performance characteristics, and then ask, what can I do with this object?
\vspace{3mm}

\textbf{Preetum Nakkiran:} I would say we actually have made progress towards understanding deep learning, it just hasn't been as formal, quantitative, provable, and prescriptive as many (myself included) had hoped. This isn't necessarily a problem --- there are many types of understanding in the natural sciences that don't satisfy these criteria either (e.g. the theory of evolution, or the germ theory of disease). 
We must thus choose our level of rigor carefully: demanding full mathematical rigor will prevent us from studying many phenomena which are real and interesting,
but which are not appropriate for provable (or even quantitative) theories.
These considerations are not new --- Turing Award laureate Herbert Simon
discussed the question of ``what is a theory?'' in the context of AI in 1995
\citep[Section 4.1]{simon1995artificial},
which remains relevant today.

There is also a long history of ``effective'' and approximate theories in the sciences, that are later refined into more fundamental and precise theories. The usual example is Newton's laws, later refined by GR and QM.
But there are many smaller, less revolutionary examples --- e.g Boyle's law was not exactly correct, and was not physically justified until much later,
and yet still captured a real phenomenon that was an important stepping stone for later theories of thermodynamics \& statistical mechanics.
We should similarly be open to starting with less rigorous, less mechanistic, less precise theories in deep learning, as a stepping stone to future theories.

Why I claim we've made progress towards understanding: If we had made no progress, deep learning would be an entirely unpredictable system to us – we would add 1 more parameter to a model, and have absolutely no idea how this new model would behave compared to the previous one. This is not the case – in fact, practitioners have very strong intuitions and heuristics about the effect of many design choices, as evidenced by their continued practical success. These intuitions and heuristics are clear evidence of partial understanding. Civilization was constructing buildings and bridges long before we had a full theory of solid mechanics.

For the second question, on whether theory would be useful to practitioners: A rich theory will almost certainly be eventually useful to practitioners, as theories in the sciences usually are. The ``eventually'' is a key qualifier, though. In the meantime, I think trying too hard to be useful to practice is a distraction for scientific theory. In this field, the goal of science is to understand the nature of learning. The goal of engineering is to build and improve learning systems. These goals are different, and not always aligned.
I do think scientists should be very aware of the developments in practice, though – they are valuable experimental points that theory must eventually capture.

Finally, a technical remark:
There are actually several definitional obstacles to having a formal, meaningful, unified theory of deep learning (as I described in my thesis \citep{PreetumThesis}). We currently lack a precise definition of the objects of study: What precisely \emph{is} a ``deep network''? We certainly cannot allow any network architecture (which would allow us to simulate a Turing Machine, and thus force us to study all of computational learning as ``deep learning''). Similarly we cannot simply enumerate common architectures, because we do not want just a ``theory of ResNets'' or a ``theory of Transformers'', but rather a theory of deep learning as a whole. 
There is no existing definition of the set of ``reasonable architectures''--- which should include all the ones that are successful in practice now, and which could be successful in the future.
There are similar obstacles to defining the ``natural distributions'' on which
we apply deep learning, and many other objects we colloquially use.
\vspace{3mm}

\phantomsection
\addcontentsline{toc}{section}{Will deep learning ever be interpretable?}
\begin{abox}
\textbf{Will deep learning ever be interpretable?} Does it need to be?  Do LLMs need to be interpretable if they can explain themselves in text?
\end{abox}
\vspace{3mm}

\textbf{Preetum Nakkiran:}
I doubt deep learning will ever be interpretable in the strong sense.
I think certain structures within networks may be interpretable,
as some existing work shows, but I expect these structures to be fairly weak.
For example, I do not expect we will ever have the following ``strong interpretability'': we will never be able to manually write an explicit program
(a Turing machine) which, without seeing any data, performs close to our strongest
trained LLMs or vision models.
However, I view the current intepretability line of work
as a valuable first step towards a ``theory'' of deep learning:
as intepretability results become more general and precise,
they essentially evolve into a theory.

Finally, we should keep in mind that these
objections are not specific to deep learning.
(Are interpolating kernel methods always interpretable?)
\vspace{3mm}

\textbf{Andrew Gordon Wilson:} Interpretability can mean many different things, and arguably the word is often taken too narrowly to mean understanding the effects of input dimensions on model outputs. But I disagree with the conventional wisdom that deep learning is particularly uninterpretable, broadly speaking. Is deep learning really much different or more mysterious than other model classes? Why? Overparametrization? Double Descent? Benign overfitting? All of these “mysterious” phenomena can be reproduced using other model classes \citep[e.g.,][]{krogh1992generalization, wilson2020bayesian, maddox2020rethinking, smith2017bayesian}. Moreover, the design decisions for many canonical deep learning architectures, such as convolutional neural networks, make a lot of intuitive sense — hierarchy, locality, translation equivariance, contrast normalization, inspired in part by our biological understanding of visual receptive fields… If that's not interpretable, what is?
\vspace{3mm}

\textbf{Alex Smola:} Given the fact that nonlinear nonparametric models are, by design, hard to understand, we can nonetheless ask the question as to whether there are simple to understand proxies (e.g. locally linear functions, local decompositions, salient components, sparse approximations, etc.), as long as we don't start believing that said proxies are the truth. For instance, clustering is generally not `true', yet it can provide a good proxy of what the data is about. See e.g. the leptograpsus and iris datasets. Clustering isn't the `truth', ditto zachary's karate club, but a pretty good proxy. 

Regarding explainability, some few years ago, SHAP and linearizations were all the rage. The idea there for explanation is that a linear function, possibly a linear function locally at f(x) would be most meaningful. But there are hundreds of alternative ways to view this, so the answer will depend on the questions just as much as on the reality. In a way, this is a beautiful example of Plato's cave. 
\vspace{3mm}

\textbf{Kyunghyun Cho:} I find interpretability as an extra feature that can be extremely useful but may not be necessary for building and deploying deep learning based systems. In fact, I believe it’s quite helpful to think of deep learning (or a system based on it) in light of what we do in healthcare and in particular drug discovery. It is quite helpful and useful to know the precise biological, chemical and physical mechanism behind any treatment, because it would help us determine the breadth of its applicability as well as some of the perils that may come together with it. However, for most of the drugs and therapeutics, we actually don’t know the exact mechanism by which they work in human bodies. How do we then prescribe these drugs and treat various conditions? We rely on randomized controlled trials (and their variants) to establish the efficacy of such therapeutics directly, bypassing the necessity of knowing the underlying mechanism. By defining a target condition and target population very carefully, we can often establish any therapeutics’ effectiveness for this particular condition-population combination, although this may require lengthy and painful trials. 
Can we also approach applying and deploying deep learning based systems in this way? I believe it is possible.

That said, interpretability is an extremely useful extra feature, as knowing the underlying mechanism allows us to more accurately guess to which population and to which conditions drugs would be effective. This can lower the effort in randomized controlled trials as well as unanticipated consequences in the future. In other words, interpretability can facilitate dramatically improving and increasing the applicability. So, yes, interpretability will be extremely helpful.

In my view, LLM’s explaining themselves has nothing to do with interpretability. Interpretability is about knowing the underlying mechanism by which certain predictions (in the context of machine learning) were made, and explanation by an LLM does not guarantee that such explanation is strongly tied to the actual underlying mechanism of the LLM. This is why interpretability often needs to go together (at least loosely) with identifiability.
\vspace{3mm}

\textbf{Melanie Mitchell:} Interpretability is one of the most important challenges for deep learning in general, and LLMs in particular. Interpretability is an essential factor in assessing robustness and trustworthiness of AI systems—it is a problem the community needs to solve. LLMs can “explain themselves” in text, but there is no guarantee that the text “explanations” they generate will have anything to do with their actual process of solving problems or making decisions. Humans are not perfect at explaining their own reasoning, but they do have significant abilities for meta-cognition—this is an important capability that LLMs currently lack. Psychology and neuroscience have provided much insight into how humans reason and make decisions. We need to discover the analogous means of teasing out the mechanisms for these capabilities in LLMs and other complex AI models.  

\vspace{3mm}

\textbf{Zachary Lipton:} This is a thorny question. Interpretability may be one of the most confused topics in all of machine learning, fraught with confusion and conflict. To begin, the word is badly overloaded. Read an interpretability paper selected at random and you’ll find representations (or insinuations) that the work is addressing ``trust'', ``insights'', ``fairness'', ``causality'', ``fairness''. Then look at what the authors actually do and you’ll be hard-pressed to tie back the method to any of these underlying motivations. Half the papers produce a set of feature important scores, describing this “importance” in cryptic ways: “what the model is looking at”, “what its internal logic depends on to make this particular prediction”.

Other papers aim to look  inside the neural network, offering vague accounts for the function of particular neurons or layers. Then there’s papers that take the “state of the art” in model interpretability, whatever that might mean, and attempt to evaluate whether these scores are actually useful in some real world human-interaction setting where they might ostensibly be deployed. Overall, this in vivo research has produced overwhelmingly negative results. And yet the methods are adopted and heavily marketed at full speed anyway.

At the heart of it all is an economic tension: there’s a general unease with trusting models in various domains. If you take the time to pick apart what precisely underlies this unease, you typically find that it’s a concern not so easily waved off. The economic role of the interpretability method is often to placate a consumer. ``Our models are trustworthy because we use {Shap/LIME/GradCAM/…}, a state of the art interpretability method!'' The entire enterprise has become polarizing, and due to reviewer bidding, this leads to a dangerous dynamic in which skeptics self-select out of the peer review cycle, leaving a cycle of affirmation.
\vspace{3mm}

\phantomsection
\addcontentsline{toc}{section}{Is deep learning moving beyond academic benchmarking?}
\begin{abox}
\textbf{Is deep learning moving beyond academic benchmarking?} As we care more and more about qualitative model properties, we care less about how many mistakes a model makes but rather what kinds of mistakes. Academic LLM benchmarks typically require single-word answers and fail to capture the complexity of real human-machine interactions. Meanwhile, vision benchmarks focus on narrow definitions of accuracy with low-diversity data. Finally, do generative AI models have more complex properties like ``common sense'' and if so, can we design benchmarks that truly capture them?
\end{abox}

\vspace{3mm}

\textbf{Alex Smola:} Benchmarking right now is a mess. Again, that's to be expected. Have a look at intelligence tests (as in, for humans) and the controversies in them.
\vspace{3mm}

\textbf{Preetum Nakkiran:} 
Just a comment (a non-answer): it is my impression that industry
has been considering more fine-grained measures of model behavior for a while,
even in ``classical'' settings such as image classification -- simply because when ML
is deployed in applications, basic QA requires testing the application
under different settings, and detailing its failure modes.
\vspace{3mm}

\textbf{Max Welling}: I agree that solid benchmarking will be the key to progress and keeping us honest. And we need to move beyond the simplistic metrics we tend to use in academia. While designing such metrics may be tricky, we have RL to integrate human feedback back into the system, and so I predict RL will start to play a key role in making these systems aligned with human values and interests. 
\vspace{3mm}

\textbf{Zachary Lipton:} I don’t know about ``moving beyond'' but I think our relationship to these benchmarks is changing rapidly. For the last ten-fifteen years, the lion’s share of activity in natural language processing consisted in academics proposing benchmarks and then doggedly targeting these benchmarks with a barrage of models. At a 1000-foot view, we lived in a world where one benchmark would summon hundreds of models. Now the script is flipped. The world of methodological innovation is moving away from single-purpose modeling efforts towards general purpose foundation models and somewhat general repertoires of tricks for how to apply them. In this new world, one-to-many becomes many-to-one. Each new foundation model is evaluated against hundreds of benchmarks. These benchmarks are seen less and less as the singular aim of a modeling effort and more as one sensor among many for addressing a more general repertoire of capabilities. 
\vspace{3mm}

\textbf{Kyunghyun Cho:}  I don’t think there’s anything special about “academic” benchmarking. That academic benchmarks typical focus on single-word answers and 0-1 errors is not that we have to do so in academia but is an indication that we have been quite lazy when it comes to evaluation. In fact, some people in e.g. natural language processing, including myself, have been working on building better, (semi-)automated evaluation protocols for language generation and others. For instance, in \citet{wang2020asking}, Alex Wang, who was a PhD student of mine back then and is now at Cohere.AI, developed an evaluation protocol for summarization systems using a chain of question-generation and question-answering models. More advanced evaluation protocols are often more difficult to develop, implement and deploy, compared to simple surface-level matching protocols, such as 0-1 error, BLEU (weighted average n-gram precision), ROUGE (weighted average n-gram recall), etc., which is why they have not been adopted more widely nor studied further. So, no, academic benchmarking is not a thing, and we can do much better on evaluation if we don’t lazily maximize the number of publications.
\vspace{3mm}

\textbf{Andrew Gordon Wilson:}  Academic benchmarks, such as ImageNet \citep{deng2009imagenet}, catalyzed the deep learning revolution. Without those benchmarks, it may have been many more years before the community embraced deep learning — perhaps it wouldn't have happened even now. There was unfortunately a lot of stigma around deep learning, and it took striking results on a benchmark the community broadly understood to start changing minds. But it's reassuring that the community did ultimately change its mind based on empirical evidence. The willingness to do so, despite many doubts, is something I deeply appreciate about the machine learning community, and we shouldn't take it for granted. Moreover, good performance on these benchmarks has served as a proxy for good performance on related problems. 

These standard benchmarks are now sufficiently well-solved that they are losing their scientific value. At this point, it is more engineering details and computation that lead to benchmark topping results than innovative ideas. 

It could be that in order to keep innovating, we need to be tailoring our approaches to more specialized problems, requiring domain expertise. If this were the case, then perhaps there is no room any longer for “standard” benchmarks. However, I believe that many even seemingly disparate real-world problems share sufficient structure \citep{goldblum2023} that there is mileage in creating fresh general purpose benchmarks. Incidentally, and perhaps counterintuitively, fears about overfitting to the test sets of academic benchmarks may also be overblown \citep{recht2019imagenet, goldblum2023}.

I'm hoping we will see continued academic benchmarking, but on fresh benchmarks.
\vspace{3mm}

\phantomsection
\addcontentsline{toc}{section}{Are transformers the way forward or are they limited?}
\begin{abox}
\textbf{Are transformers the way forward or are they limited?} What are the fundamental limitations of today’s popular paradigms? Do we need a post-deep-learning paradigm?  Will future improvements come primarily from hardware improvements or advancements in fundamental methods?
\end{abox}

\vspace{3mm}

\textbf{Alex Smola:} Are transformers the answer? Obviously not. Just like the Fourier transform isn't the `answer', but they're by now baked into most deep learning silicon with great acceleration. So we're stuck with them unless someone figures out another function class that can either significantly beat transformers or that can beat them somewhat and use the same types of acceleration. It might take another 5 years for this to happen, maybe 10 years due to the inertia of deployed hardware. 
\vspace{3mm}

\textbf{Max Welling:} We find ourselves currently in a scaling paradigm, where most of the improvement seems to come from scaling to more data and bigger models. But I predict that will saturate at some point and we will revert back to modeling. This can happen either because the data is exhausted (or in some domains not available in large amounts), or dealing with very large models becomes too expensive. Machine learning really is a balancing act between inductive bias and data, and as such I think at some point we will find other architectures beyond transformers that will scale and have smarter inductive biases embedded in them.
\vspace{3mm}

\textbf{Kyunghyun Cho:} Transformers were definitely the way forward. In fact, transformers are the culmination of the best practices we’ve discovered over the past three decades; stochastic gradient descent, linear shortcut connections, attention, rectified linear units and normalization. It is almost unsurprising how well Transformers work. It is in fact surprising that people are surprised how well Transformers work. 

I don’t know what the future will be nor should be. I believe in the diversity of scientific topics and directions pursued by scientists. So, the answer is ``all of the above''.
\vspace{3mm}

\textbf{Zachary Lipton:} Transformers have been a remarkably powerful architecture, but my overall sense is that the role of architectures has been a bit overstated. It’s not clear to me that we couldn’t achieve similar feats by putting a similarly massive push behind something that looked more like a traditional RNN. Of course, Transformers have computational advantages. But I think the real action now has been on the fronts of massively parallel computation (e.g. training GPT4 and PaLM on thousands of pods of linked GPU/TPU nodes), web-scale data, and a repertoire of pre-training, instruction fine-tuning, task fine-tuning, and in-context learning, and evaluation routines that have moved us into new terrain.
\vspace{3mm}

\textbf{Preetum Nakkiran:} I don't think transformers are the end-all-be-all, for at least two reasons.
First, because it is a very strong statement to claim that a certain learning method is globally optimal over a broad family of distribution, metrics, and design constraints.
Second, because engineered systems are typically always being improved – we're still improving on the automobile, for example. An engineered system is unlikely to be optimal, unless it is solving a very well-defined problem, and hitting fundamental physical limits (e.g. Carnot engines).
I am optimistic that a better understanding of which parts of the transformer architecture are crucial, and why, will lead us to better architectures which directly exploit these advantages.
\vspace{3mm}

\textbf{Andrew Gordon Wilson:} 
 They are the way forward, and they are limited. Transformers show us it is possible to develop relatively general purpose algorithms, capable of solving problems in a remarkably wide array of domains, in contrast to what is often implied from the “no free lunch theorems” \citep{goldblum2023}. But, at least in their current form, they are not the “final stage of evolution” for learning algorithms. For example, they are relatively data hungry. Perhaps research showing how to include soft priors, without constraining expressivity, will help resolve this issue \citep{finzi2021residual}. Integrating planning will also be an important future direction for these models.  Multimodal learning is also the future --- I'll stick my neck out and say in five years a majority of applied papers at NeurIPS, ICML, and ICLR will be considering multimodal datasets, and we still stop thinking about vision and NLP so much as distinct problems. While transformers can handle multimodal data, we will likely develop non-trivially different architectures to accommodate new modalities and fusion of modalities. For LLMs, we may also move away from autoregressive models. I also don't think text alone is the path towards general purpose intelligence.

\vspace{3mm}

\textbf{Melanie Mitchell:} Whether transformers are the “way forward” depends on what the goal is.  For something like “human-level intelligence,” it’s clear that the architecture and learning methods of today’s transformers have many differences from biological brains and learning methods.  Here are some central aspects of human brains/minds and learning methods that are missing (or nascent) in transformers. 

\begin{itemize}
\item Episodic memory — capability of long-term storage (and context-sensitive retrieval) of experiences during one’s lifetime

\item Metacognition — being able to report on and reason about one’s own thinking.

\item Ability to dynamically simulate / imagine possible physical and social situations, reason about them, make predictions

\item Active interaction with the world / environment — choosing what to learn from, what interventions to make to test predictions and adjust internal models

\item Sensory systems, multimodal integration

\item Motor systems, integration of brain and body

\item These are just some examples; there are many other salient differences between biological and (current) artificial intelligence. 
\end{itemize}

\vspace{3mm}

\phantomsection
\addcontentsline{toc}{section}{Are academics screwed?}
\begin{abox}
\textbf{Are academics screwed?} 
\end{abox}

\vspace{3mm}

\textbf{Kyunghyun Cho:} No.
\vspace{3mm}

\textbf{Preetum Nakkiran:} Depends on how you define academic \raisebox{-.4ex}{\twemoji{wink}}
\vspace{3mm}

\textbf{Alex Smola:} Obviously not. A lot of things are possible in academia that are difficult in industry. Taking risks, taking risks with datasets, and also, asking questions such as explainability and interpretability. Ditto optimizing algorithms. That said, deep learning research is starting to look a lot more like the work in biological wetlabs or experimental physics. All of them require significant capital investment to work. Once upon a time, you only needed pen and pencil to make something work. This is only less the case by now.
\vspace{3mm}

\textbf{Max Welling:} Not at all. But they may be playing a different game. Perhaps the engineering related to scaling is best left to industry, but there are very many interesting fundamental, theoretical, exploratory problems that can be solved better in an academic setting. For instance, to really improve AI for Science, we need engineers, machine learning researchers and scientists (e.g. physicists, chemists, biologists). The best place where the crème de la crème of these disciplines comes together is academia. I would say, if you are in academia, pick your problem wisely!
\vspace{3mm}

\textbf{Andrew Gordon Wilson:}  
No. For the most part, industry and universities (or ``academia'', more broadly) play complementary rather than competing roles. 

It's true that the approach to building bigger models trained on more data has been fruitful, is more suited to industry resources, and not particularly academic. 

But suppose you are modelling a scientific problem where you know you want to constrain your neural network to respect a symmetry (e.g., rotation). Perhaps you want to study molecules, which have no canonical orientation or coordinate system. That's not a question that should be solved by brute force scale. It's answered by careful consideration, and mathematics \citep{cohen2016group}. Indeed, many recent machine learning advances have been driven by academia --- even in areas most crucial to real-world adoption, such as robustness to spurious correlations and distribution shifts, uncertainty representation, and online decision making \citep[e.g.,][]{sagawa2019distributionally, kirichenko2022, arjovsky2019invariant, izmailov2021bma, liu2021just, wilson2020bayesian, snoek2012practical, balandat2020botorch}. Moreover, universities are well-suited to open-ended long-term fundamental research that would be hard to ``sell'' to an industry manager or align with short-term company priorities. 

There will always be an important role for fundamental research. To come full circle, surely scientific discovery is an academic endeavour.

\textbf{Zachary Lipton:} In a holistic sense, obviously not! Or if we are screwed, the threat may be as much from online lecture material and ChatGPT attacking the necessity  and undermining the integrity of coursework as it is from our ability to compete in the research landscape. There are so many profound problems for scholars to work on and while building the biggest foundation models may be a fundamentally important activity and is doubtless driving a lot of the action, it’s (i) just one activity and (ii) arguably not the research line most compatible with the unique advantages of academia. I expect the frontiers of research on the statistical foundations of ML, on causal inference and related decision-making disciplines, etc. to remain in academia. If you take a narrow view of “academics” as leading the industrial edge of deep learning then sure academics might be screwed, and those of us with dual affiliations might be half/screwed. The situation, at worst, could be likened to the status of information retrieval research. Prior to Google, there was a time where the most exciting work on web search took place in the hall of academia, and eventually we became somewhat obsolete, keepers of a public body of work that lagged grotesquely far behind the vanguard of industry. But that was OK. Search became (mostly) the province of industry, and academics focused on other exciting directions (like neural networks). A primary strength of academia is providing a lane to focus intensely on problems that don’t have the clear path and immediate economic impact required by industrial R\&D. Academics as a whole aren’t screwed, but many might need to refocus given the shifting landscape (or move into industry).
\vspace{3mm}

\hspace{-1mm}\raisebox{-.4ex}{\emojigpt}\textbf{ChatGPT-3.5:} Academics aren't screwed, they're just upgrading their syllabus from ``Old McDonald Had a Data Set'' to ``GPT-3.5 Ways to Impress Your Robot Overlords.'' \raisebox{-.4ex}{\twemoji{wink}}

\hspace{-1mm}\raisebox{-.4ex}{\emojigpt}\textbf{ChatGPT-4:}
Oh, such melodrama. ``Screwed,'' you say? Like suddenly, all academic researchers are going to disappear into a void, swallowed whole by the corporate behemoth? What an imaginative dystopia you've conjured up!

Let me dispel that illusion for you.
Yes, it's true that the industrial tech giants have been flexing their muscles with large-scale language models. It's also true that they have buckets of money to grease the wheels of their research. But let's not pretend that this spells the end for academic machine learning researchers.

Academic research isn't just about being the first to train a massive language model or the ability to throw wads of cash at a problem. It's about creating knowledge, asking fundamental questions, exploring new areas not yet colonized by the profit-oriented industrial labs.

While industry labs are racing to build bigger and glossier models, academic researchers play a vital role in providing the much-needed critical lens, asking questions about ethics, interpretability, robustness, and social implications. They also innovate in other areas where resource constraints actually foster creativity and inventiveness.
So, screwed? Hardly. Challenged? Definitely. But adversity, my dear inquirer, often sparks the most brilliant flames of creativity and resilience. So let's not start drafting the obituary for academic machine learning research just yet. It's got a lot more fuel left in the tank. If anything, this could be the wakeup call needed to stop the homogenization of research and encourage a more diverse ecosystem of ideas.

\section*{Editors and Contributors}

\textbf{Editors:}
\begin{itemize}
    \item \href{http://goldblum.github.io}{Micah Goldblum} - \textit{New York University}, \url{goldblum@nyu.edu}
    \item \href{http://tensorlab.cms.caltech.edu/users/anima/}{Anima Anandkumar} - \textit{California Institute of Technology, NVIDIA}, \url{arangelf@caltech.edu}
    \item \href{https://richb.rice.edu/}{Richard Baraniuk} - \textit{Rice University}, \url{richb@rice.edu}
    \item \href{https://www.cs.umd.edu/~tomg/}{Tom Goldstein} - \textit{University of Maryland}, \url{tomg@umd.edu}
\end{itemize}
\textbf{Contributors:}
\begin{itemize}
    \item \href{https://kyunghyuncho.me/}{Kyunghyun Cho} - \textit{New York University, Prescient Design}, \url{kyunghyun.cho@nyu.edu}
    \item \href{https://www.zacharylipton.com/}{Zachary Lipton} - \textit{Carnegie Mellon University}, \url{zlipton@cmu.edu}
    \item \href{https://melaniemitchell.me/}{Melanie Mitchell} - \textit{Santa Fe Institute}, \url{mm@santafe.edu}
    \item \href{https://preetum.nakkiran.org/}{Preetum Nakkiran} - \textit{Apple}, \url{preetum@nakkiran.org}
    \item \href{https://alex.smola.org/}{Alex Smola} - \textit{Boson AI}, \url{alex@smola.org}
    \item \href{https://staff.fnwi.uva.nl/m.welling/}{Max Welling} - \textit{University of Amsterdam}, \url{M.Welling@uva.nl}
    \item \href{https://cims.nyu.edu/~andrewgw/}{Andrew Gordon Wilson} - \textit{New York University}, \url{andrewgw@cims.nyu.edu}
    \item \href{https://chat.openai.com/}{ChatGPT}\raisebox{-.4ex}{\emojigpt} - \textit{OpenAI}
\end{itemize}

\bibliography{mbibnewest}

\end{document}